 \documentclass[pmlr,twocolumn,10pt]{jmlr} 

\usepackage{booktabs}
\usepackage{siunitx}
\usepackage[most]{tcolorbox}
\usepackage{amssymb}
\usepackage[switch]{lineno}

\newcommand{\equal}[1]{{\hypersetup{linkcolor=black}\thanks{#1}}}

\newtcolorbox{simplebox}{
  colback=black!5!white,     
  colframe=black!25!white,   
  boxrule=0.5pt,              
  arc=1mm,                    
  left=3mm, right=3mm, top=3mm, bottom=3mm,
  fontupper=\small
}

\newcommand{\hangitem}[2][14pt]{%
  {\par\hangindent=#1 \hangafter=0 \noindent #2\par}%
}
\usepackage{xcolor}
\usepackage[table]{xcolor}
\usepackage{caption}
\usepackage{titlesec}
\usepackage{tikz}
\usetikzlibrary{positioning,fit,shapes,arrows.meta}
\usepackage{subcaption}

\titlespacing*{\section}{0pt}{5pt}{2pt}
\titlespacing*{\subsection}{3pt}{3pt}{0pt}

\definecolor{backgroundblue}{HTML}{4F81BD}
\definecolor{guidelineorange}{HTML}{E46C0A}

\theorembodyfont{\upshape}
\theoremheaderfont{\scshape}
\theorempostheader{:}
\theoremsep{\newline}

\jmlrvolume{297}
\jmlryear{2025}
\jmlrworkshop{Machine Learning for Health (ML4H) 2025} 

\title[Dialogue to Question Generation for EBM]{Dialogue to Question Generation for Evidence-based \\ Medical Guideline Agent Development}

\author{%
\Name{Zongliang Ji}\equal{Equal contribution. Work done while interning at Google.}\textsuperscript{1,2,3} \Email{jerryji@cs.toronto.edu} \\
\Name{Ziyang Zhang}\footnotemark[1]\textsuperscript{1,4} \Email{ziyang.zhang2@emory.edu} \\
\Name{Xincheng Tan}\textsuperscript{1} \Email{caratan@google.com} \\
\Name{Matthew Thompson}\textsuperscript{1} \Email{mthomp@google.com} \\
\Name{Anna Goldenberg}\textsuperscript{2, 3} \Email{anna.goldenberg@utoronto.ca} \\
\Name{Carl Yang}\textsuperscript{4} \Email{j.carlyang@emory.edu} \\
\Name{Rahul G. Krishnan}\textsuperscript{2, 3} \Email{rahulgk@cs.toronto.edu} \\
\Name{Fan Zhang}\textsuperscript{1} \Email{zhanfan@google.com} \\
\addr \textsuperscript{1} Google Research\\
\addr \textsuperscript{2} University of Toronto, Canada\\
\addr \textsuperscript{3} Vector Institute, Canada \\
\addr \textsuperscript{4} Emory University, USA \\
}

\begin{document}

\maketitle

\begin{abstract}
Evidence-based medicine (EBM) is central to high-quality care, but remains difficult to implement in fast-paced primary care settings. Physicians face short consultations, increasing patient loads, and lengthy guideline documents that are impractical to consult in real time. To address this gap, we investigate the feasibility of using large language models (LLMs) as ambient assistants that surface targeted, evidence-based questions during physician–patient encounters. Our study focuses on question generation rather than question answering, with the aim of scaffolding physician reasoning and integrating guideline-based practice into brief consultations. We implemented two prompting strategies, a zero-shot baseline and a multi-stage reasoning variant, using Gemini 2.5 as the backbone model. We evaluated on a benchmark of 80 de-identified transcripts from real clinical encounters, with six experienced physicians contributing over 90 hours of structured review. Results indicate that while general-purpose LLMs are not yet fully reliable, they can produce clinically meaningful and guideline-relevant questions, suggesting significant potential to reduce cognitive burden and make EBM more actionable at the point of care.
\end{abstract}

\begin{keywords}
LLMs, Evidence-based medicine, Clinician-facing tools, Clinical Decision Support System
\end{keywords}

\paragraph*{Data and Code Availability}
The data we use is not entirely shareable as the dataset is preparatory. 
We share some generated evidence-based medical questions in the shared code base.
The prompts and partial data used for this study is contained in this GitHub repository \footnote{https://github.com/Jerryji007/Dialogue2Questions-ML4H2025}. 



\paragraph*{Institutional Review Board (IRB)}
Our study does not require IRB.


\section{Introduction}
\label{sec:intro}

\begin{figure*}[ht]
    \centering
    \includegraphics[width=\linewidth]{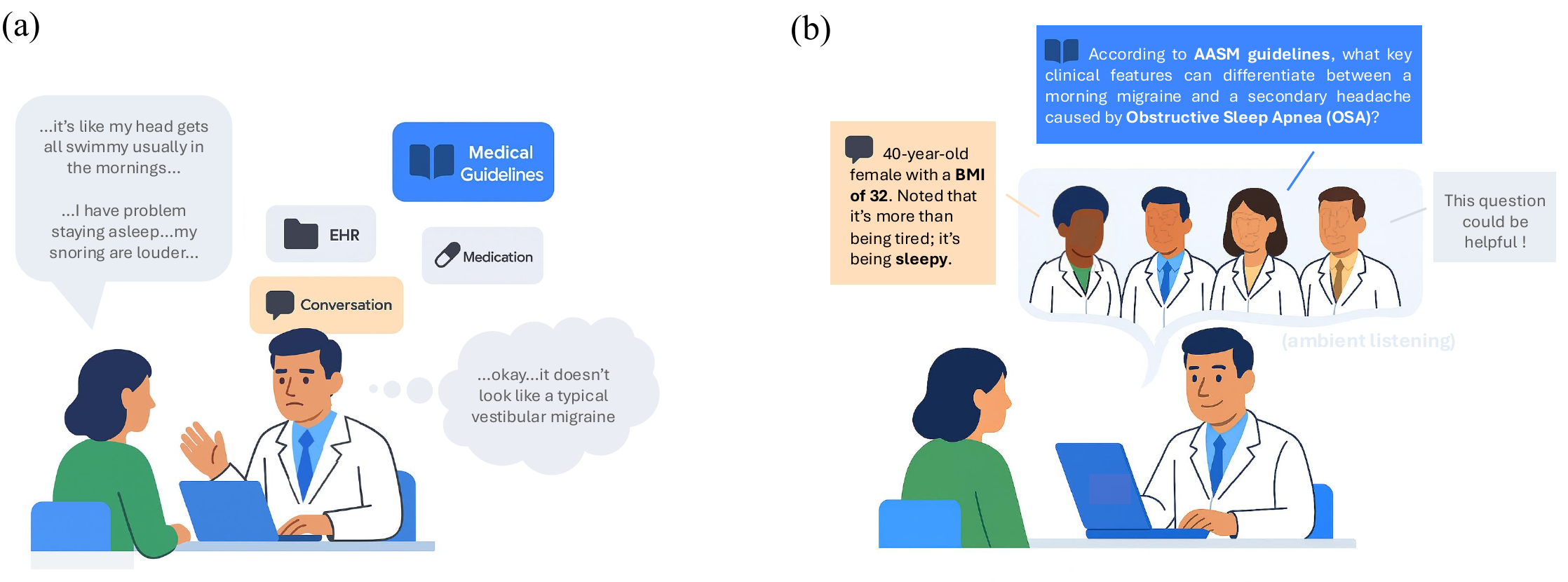}
    \caption{\textbf{Illustration of an outpatient visit.}
A 40-year-old female with a high BMI presents with headache and fatigue.
(a) In a routine encounter, the PCP simultaneously juggles the patient’s narrative, EHR, prior comorbidities, and guidelines, while considering differential diagnoses such as vestibular migraine.
(b) With intelligent support, the PCP receives context-aware, evidence-based information that help structure reasoning without interrupting patient communication.}
    \label{fig:intro}
\end{figure*}

Evidence-based medicine (EBM) is central to the delivery of high-quality care. Its principles are operationalized through clinical guidelines, which summarize research evidence into structured recommendations designed to support everyday clinical decisions \citep{panteli2019clinical}. Despite their importance, primary care physicians (PCPs) frequently struggle to apply guidelines in practice. Consultation times are short, often less than 15 minutes in the United States \citep{neprash2021measuring}, and the format of guidelines as lengthy reference documents requiring precise keyword searches makes them impractical to use during real-world encounters \citep{pondicherry2023designing}. As a result, guideline-based decision-making often remains more aspirational than achievable at the point of care.

Consider the scenario in Figure~\ref{fig:intro}a. A PCP must listen, extract key clinical details from the narrative, review the electronic health record (EHR), and begin differential diagnosis. Several guidelines may apply, but cross referencing them in real time creates substantial cognitive load. Figure~\ref{fig:intro}b envisions a silent virtual expert that listens to the dialogue and raises targeted, evidence based questions, then uses retrieval to answer them by synthesizing guideline content. The ability to surface the right questions is not merely preparatory but decisive: it defines the retrieval space, shapes clinical reasoning, and directly reduces cognitive load when applying guidelines under time pressure. Question generation therefore constitutes a high-impact problem on its own, even before answer synthesis is integrated. This paper formalizes and studies the question generation in depth in the primary care outpatient setting.

Previous efforts to assist clinicians with guideline use include static mobile applications \citep{mitchell2020development}, electronic health record order sets \citep{bates2003ten}, and computerized clinical decision support systems (CDSS) \citep{sutton2020overview}. While these approaches improved accessibility, they often lacked context-awareness and imposed rigid workflows. More recent developments have shifted toward AI-driven assistants and large language models (LLMs), which allow clinicians to query guidelines in natural language \citep{sagheb2022artificial,lichtner2023automated} or receive retrieval-augmented responses that make guideline content more actionable at the point of care \citep{oniani2024enhancing,ferber2024gpt}. Despite these advances, few systems have been explicitly designed or evaluated for proactive, real-time question generation during visits \citep{fast2024autonomous,hager2024evaluation}.

In this work, we investigate the feasibility of using LLMs as ambient question generators that support physicians in practicing evidence-based medicine. Specifically, our contributions are as follows:

\begin{itemize}
\item We identify short primary care visits as a central barrier to practicing EBM and propose, for the first time, an ambient LLM that listens to the encounter, surfaces guideline oriented questions, and later retrieves and synthesizes answers. In this paper we address the first step by generating EBM questions from dialogue, instantiated with two prompting strategies: zero shot and multi stage.
\item We curate an evaluation benchmark consisting of 80 de-identified real-world physician–patient dialogue transcripts, each presented at three different truncation lengths. To systematically assess question quality, we design five complementary metrics across different aspects and experimental tasks for evaluation.
\item We conduct an extensive human evaluation with six experienced physicians, who collectively devoted more than 90 hours to assessing the generated questions. In parallel, we set up an LLM-as-judge evaluation for comparison. These experiments yield insightful findings on the necessity of such systems, the plausibility of an ambient assistant, the comparison of prompting methods, the variation across question types, and the promises \& limitations of automated evaluation.
\end{itemize}


%

\section{Related Work}
\textbf{Challenges in Practicing Evidence-Based Medicine }
Primary care physicians (PCPs) face longstanding structural barriers to applying evidence-based medicine (EBM). Time studies estimate that a solo PCP would require 7-18 hours per day for preventive and chronic care alone \citep{Yarnall2003,Ostbye2005}, and nearly 27 hours per day to fully implement guideline recommendations for a typical patient panel \citep{Skandari2023}. These demands intersect with a persistent workforce shortage, with the U.S. projected to lack over 50,000 PCPs by 2025 \citep{Petterson2012}. Consequently, many evidence-supported interventions are omitted in practice, with lack of time, difficulty accessing up-to-date guidance, and limited EBM training cited as key barriers \citep{Zwolsman2012}. At the point of care, roughly half of physicians’ clinical questions remain unanswered due to workflow constraints \citep{DelFiol2014}. Classic clinical decision support systems (CDSS) have shown benefits in narrow contexts \citep{McGinn2013}, but systematic reviews highlight mixed patient-level outcomes and barriers such as poor usability and alert fatigue \citep{Garg2005,Kawamoto2005,Vasey2021}.

\textbf{LLMs in Healthcare and Clinical Support }
Recent advances in LLMs offer new opportunities for ambient, real-time support. LLMs have achieved expert-level accuracy on medical QA tasks \citep{Singhal2025} and produced patient-facing answers judged more empathetic and higher quality than physicians’ responses \citep{Ayers2023}. They have also been applied to clinical summarization, sometimes outperforming clinicians \citep{FraileNavarro2025, VanVeen2024}. Retrieval-augmented approaches enhance grounding in guidelines \citep{Shi2024}. Emerging systems like MediQ \citep{Li2024}, HealthQ \citep{Wang2025}, and FollowupQ \citep{Gatto2025} explore proactive question generation.  
Large-scale conversational agents (e.g., AMIE, KERAP) demonstrate diagnostic accuracy comparable to PCPs \citep{Tu2025, xie2025kerap}. Early trials show that GPT-4-based support can improve physician decision-making without worsening bias \citep{Goh2025a,Goh2025b}, though limitations remain in nuanced clinical reasoning \citep{Hager2024}. Overall, LLMs show promise as ambient assistants to reduce PCP cognitive load and integrate guideline-based reasoning into constrained visits.

Across prior work, EBM remains necessary yet underutilized in routine care, and existing tools have not effectively reduced the friction of guideline use during real encounters. Meanwhile, LLMs show promise in clinical NLP tasks. This work is, to our knowledge, the first to systematically assess whether an LLM can serve clinicians by proactively generating evidence-oriented, guideline-targeting questions during encounters, with structured, multi-physician human evaluation.

\section{Problem Definition}

Our long-term vision is to develop an ambient assistant that silently observes the primary care consultation and raises clinically meaningful questions grounded in EBM. 

A downstream component, which is not part of this work, could then \textbf{retrieve and summarize the relevant guideline content} to support the physician. In this paper, we focus exclusively on the \textbf{generation of targeted, evidence-based questions} during the encounter.

We assume access to two main inputs. The first is the patient health record (PHR), represented by a structured intake questionnaire and basic clinical background, denoted as $x_{phr}$. The second is the dialogue between the patient and the physician, denoted as $x_{dlg}$. 
To approximate a realistic ambient setting, we model truncated dialogue contexts with a ratio parameter $r \in \{0.3, 0.7, 1.0\}$. The truncated dialogue is defined as
\[
x_{dlg}^{(r)} = \text{First } r \times |x_{dlg}| \text{ tokens of the dialogue}.
\]

This setting reflects two assumptions: the assistant may only have partial knowledge of the visit at a given time, and it may need to surface questions before the encounter is completed.

Given the pair $(x_{phr}, x_{dlg}^{(r)})$, the system generates a small set of questions: \(Q = \{q_1, q_2, q_3\}\). We constrain the output to exactly three questions in order to maintain readability and align with the time-limited nature of outpatient consultations. The questions should not be trivial facts that a physician is expected to recall from memory; instead, they should require reference to guidelines or evidence-based resources. 

\begin{figure*}[ht]
    \centering
    \includegraphics[width=\linewidth]{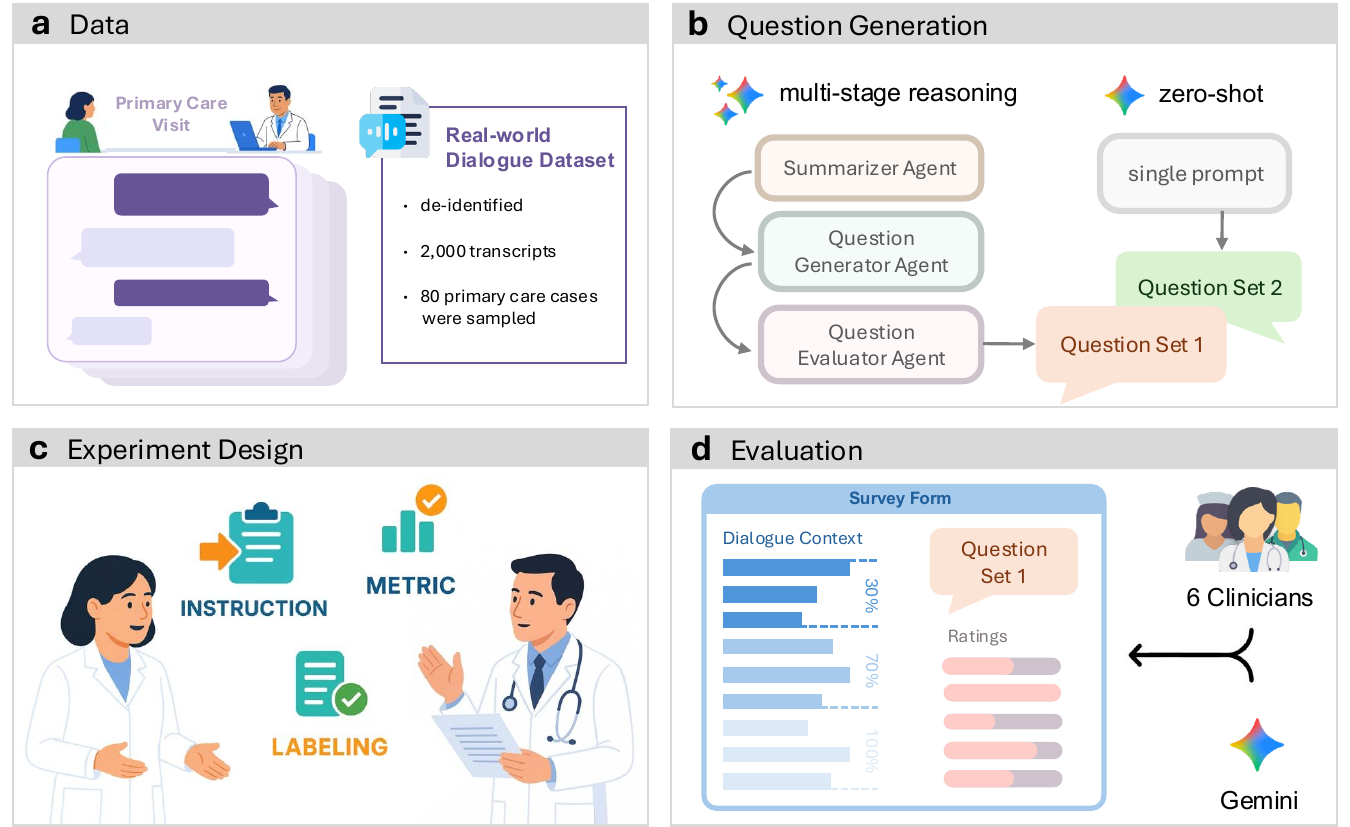}
    \caption{\textbf{Overview of our work.} (a) We sample 80 primary care cases from 2,000 real-world physician-patient dialogue transcripts. (b) We develop two methods to generate evidence-based medical questions from dialogue using Gemini 2.5. (c) We conduct a pilot study with more than 10 internal clinical experts regarding the evaluation metrics, pilot labeling, and rationale of designing the experiments. (d) We perform auto evaluation with LLM and human evaluation on 80 cases with 6 experienced clinicians.
    }
    \label{fig:method}
\end{figure*}

\section{Method}\label{sec:method}

\subsection{Backbone Model}

The Gemini family of models has demonstrated strong performance, scalability, and flexibility in the fields of medicine and healthcare \citep{saab2024capabilities, vedadi2025towards, saab2025advancing}. In this work, we use Google's Gemini 2.5~\citep{comanici2025gemini} 
as our backbone model . Gemini 2.5 contains Flash and Pro variants. For our Summarizer Agent, we use the Flash version for efficiency. For Question Generator Agent and Question Evaluator Agent, we use the Pro version for better quality. 

\subsection{Baseline: A Zero-shot Setting}

As a baseline, we consider a simple zero-shot setting where the LLM directly takes the dialogue \(x_{\text{dlg}}\) and the patient questionnaire \(x_{\text{phr}}\) as input, to generate three candidate questions, without any few-shot exemplars or structured reasoning steps (Figure~\ref{fig:method}b). The prompt is presented in Appendix~\ref{apd:first}.

\subsection{Multi-stage Reasoning Framework}\label{sec:method:proposed}

\textit{\textbf{Summarizer Agent.}} The initial and most critical step in building an effective medical guidelines agent is to ensure all clinical information is accurately captured from the dialogue context, but this verbal communication is often unstructured and prone to information loss \citep{lange2024lost}. Additionally, in real-world conversation, the patient and the doctor are not consistently communicating in professional ways. In many cases, their exchanges include low-density information or clinically irrelevant content, such as greetings or small talks, which does not directly contribute to identifying clinical conditions.


We design a \textit{Summarizer Agent} to extract the essential information from such a verbose conversation and generate structured clinical documentation.
It acts as the first stage of our multi-stage reasoning framework. Its objective is to listen to the dialogue, given the PHR, and generate a structured clinical summary (full prompt in Appendix~\ref{apd:first}). Formally, the input to the LLM agent is a pair $(x_{phr}, x_{dlg}^{(r)})$. The output is a structured summary $s$, represented as a schema of clinical slots:
\begin{equation}
s = \{(k_1, v_1), (k_2, v_2), \dots, (k_m, v_m)\},
\end{equation}
where each key $k_i$ corresponds to a predefined clinical field such as \textit{\textbf{chief complaint}}, \textit{\textbf{history of present illness}}, \textit{\textbf{medication and past history}}, \textit{\textbf{objective findings}}, \textit{\textbf{assessment}}, and \textit{\textbf{plan}}. These fields draw on common clinical documentation practices such as the SOAP note structure \citep{podder2021soap}, HL7 CDA \citep{dolin2001hl7}, and FHIR \citep{ayaz2021fast}, which adopt similar section-based structures to ensure completeness and interoperability.

\vspace*{2pt}
\noindent \textit{\textbf{Question Generator Agent.}} Through consultation with more than 10 internal clinical experts\footnote{These internal collaborators are different from the clinicians participated in the later evaluation study.}, we identified scenarios where physicians typically consult guidelines: complex diagnoses, treatment initiation or adjustment, preventive screening, comorbidity management, and referral decisions. To reflect U.S. primary care practice, we also consider the distribution of common visit types and diagnoses~\citep{ashman2023characteristics}. A good generated question should be helpful, evidence-based, and insightful. Here is an example:

\hangitem{\hspace*{-4pt}\textit{\small{``\textcolor{backgroundblue}{A 40-year-old female with a history of migraines presents with new-onset morning headaches and fatigue.} According to the \textcolor{guidelineorange}{American Academy of Sleep Medicine clinical practice guidelines}, what key clinical features can help differentiate between a primary headache disorder and a headache caused by a strong clinical suspicion of Obstructive Sleep Apnea?"}} }

\noindent We follow a few-shot setting to ensure that the generated questions align with the high-quality \textit{golden standard} questions verified by our experts. In particular, each question first introduces the necessary personal health background and key conditions (\textcolor{backgroundblue}{in blue}) and then suggests a very specific guideline (\textcolor{guidelineorange}{in orange}), acting as a \textit{mental prompt} that exists only implicitly in the physician’s mind, yet still consumes effort to think through or express.  The LLM is prompted to analyze the given summary $s$ generated by the \textit{Summarizer Agent}, remain strictly grounded in the provided context to avoid hallucinations, and output exactly ten diverse questions across categories such as \textbf{\textit{medication adjustment}}, \textbf{\textit{ordering tests}}, \textbf{\textit{medication details}}, \textbf{\textit{diagnosis}}, \textbf{\textit{follow-up}}, and \textbf{\textit{counseling}}, with two expert-verified examples included in the prompt (see Appendix~\ref{apd:first}).

\vspace*{2pt}
\noindent \textit{\textbf{Question Evaluator Agent.}} Not all ten generated questions are shown to physicians. To select the top-3 questions in terms of their quality, we run a candidates evaluation in which the \textit{Question Evaluator Agent} rates each question (1.0–5.0) on seven predefined criteria (see their definition and prompt in Appendix~\ref{apd:first}). Empirically, LLM evaluators often give uniformly high scores because the questions were generated by a state-of-the-art model, yet meaningful comparison requires more differentiated scoring. We employ a chain-of-thought (CoT) procedure inspired by \citet{liu2025proactive}: for each candidate, the LLM briefly reasons about the strongest pros and cons (grounded in the seven criteria), then assigns scores for all criteria before proceeding to the next question. Let $c_{ij}$ be the score for the $i$-th question under the $j$-th criterion; we select the top-3 by the mean score across criteria:
\begin{equation}
\small{
\text{Selected Questions} 
=\\
\operatorname*{arg\,max}_{\substack{Q \subseteq \{1,\dots,10\}\\|Q|=3}}
\;\; \sum_{i \in Q} \left( \frac{1}{7} \sum_{j=1}^7 c_{ij} \right).}
\end{equation}

\section{Experiment Design}

\subsection{Dataset and Preprocessing}

We use a large de-identified dataset of medical dialogues previously adopted in AMIE \citep{Tu2025}. The corpus includes 2,000 U.S. clinical visit transcripts spanning various specialties and conditions (Figure~\ref{fig:method}a). For this study, we restricted to primary care, family medicine, and internal medicine, yielding 899 cases. After filtering the top and bottom 5\% by length, 810 remained, with an average of 1,639 words and 150 turns. From these, we sampled 80 diverse encounters based on their quality with complete patient records as our evaluation set.

\subsection{Human Evaluation Design}

We conducted a structured human evaluation with six practicing clinicians, focusing on how generated questions could support evidence-based medicine in real consultations. To ensure clarity and consistency, we worked with an experienced collaborator to draft detailed rater instructions (Figure~\ref{fig:method}c, Appendix~\ref{apd:instructions}). These instructions emphasized that the system acts as an ``evidence-based guideline assistant,'' raising questions that clinicians might normally consult guidelines to answer.

Each evaluation consisted of a patient health record, a truncated dialogue ($r \in \{0.3,0.7,1.0\}$), and a set of three generated questions (Figure~\ref{fig:method}d). For every case, clinicians compared two model variants (zero-shot baseline and multi-stage framework), producing six total evaluations per case. Annotators assessed each set on five dimensions using a 7-point Likert scale: \textit{Relevance}, \textit{Guideline Navigation}, \textit{Thought Alignment}, \textit{Non-Redundancy}, and \textit{Usefulness} (Table~\ref{tab:metric}). These criteria were chosen after iterative discussions to reflect real-world value for PCPs. Beyond rating, annotators selected the single most useful question or indicated that no question was necessary, with the option to propose an alternative. 

\vspace*{3pt}

\begin{table}[h] 
 \centering 
 \small
 \renewcommand{\arraystretch}{1.5}
 
 \definecolor{lightgray}{gray}{0.94}
 \rowcolors{2}{}{lightgray}
\footnotesize
 \begin{tabular}{p{1.5cm} p{5.8cm}} 
 \toprule 
 \textbf{Metric} & \textbf{Statement} \\ 
 \midrule 
 Relevance & The questions are relevant and highlight insightful aspects of the case. \\ 
 Guideline Navigation & The questions guide me toward which specific guidelines for evidence I should consult. \\ 
 Thought Alignment & The questions align with my own clinical reasoning or thought process, without challenging my judgment. \\ 
 Non-Redundancy & The questions are not redundant and will not impose a cognitive burden during the visit. \\ 
 Usefulness & I would find these questions genuinely helpful for saving time and improving my daily workflow.\\ 
 \bottomrule 
 \end{tabular} 
 \caption{\textbf{Five metrics for evaluation.}} 
 \label{tab:metric} 
\end{table}

The evaluation was run as a survey over 80 cases\footnote{The clinician evaluation interface is shown in Appendix \ref{apd:second}.}. Each case had a fillable questionnaire containing the patient health record, the truncated dialogue, and both model outputs. Clinicians used a central progress list linking to each questionnaire, which allowed annotators and the study team to track completion. This survey format minimized technical friction; pilot timing was 25 to 30 minutes per case, for a total of more than 90 hours of expert review.

We also supplied the same rater instructions to Gemini 2.5 Pro and used it as an automated evaluator on the same cases; we later compare these scores with the human ratings.






\newsavebox{\imagebox}

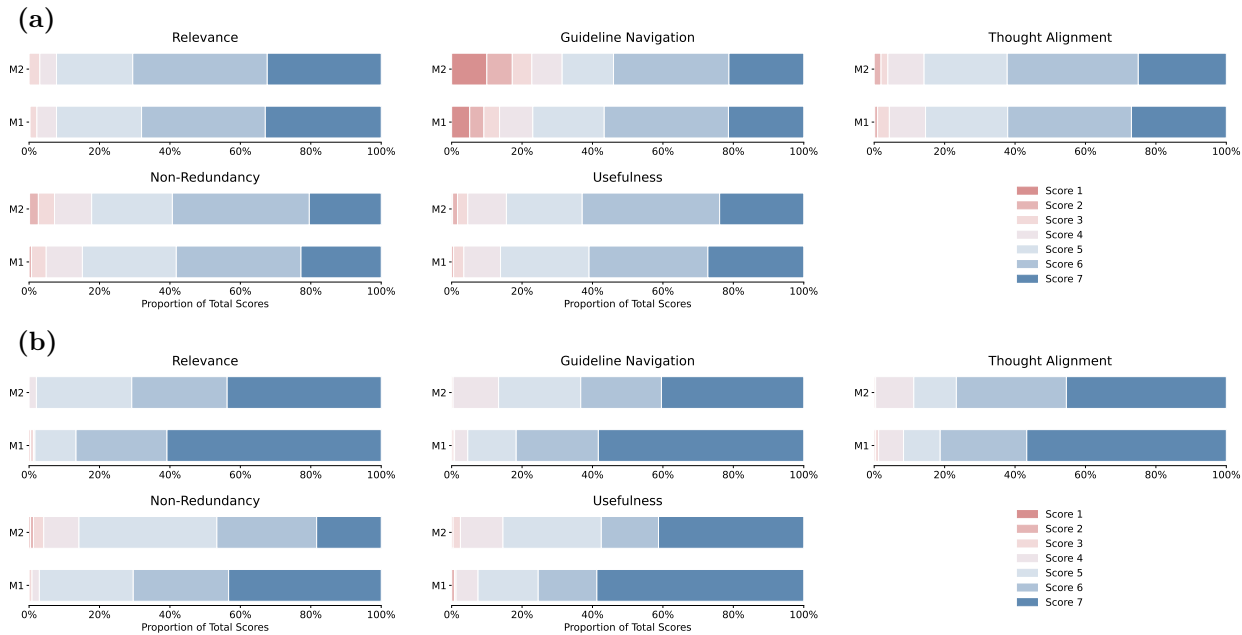
\begin{figure*}[ht]
    \centering
    \setlength{\unitlength}{1pt}
    \begin{tabular}{c}
        \sbox{\imagebox}{\includegraphics[width=\linewidth]{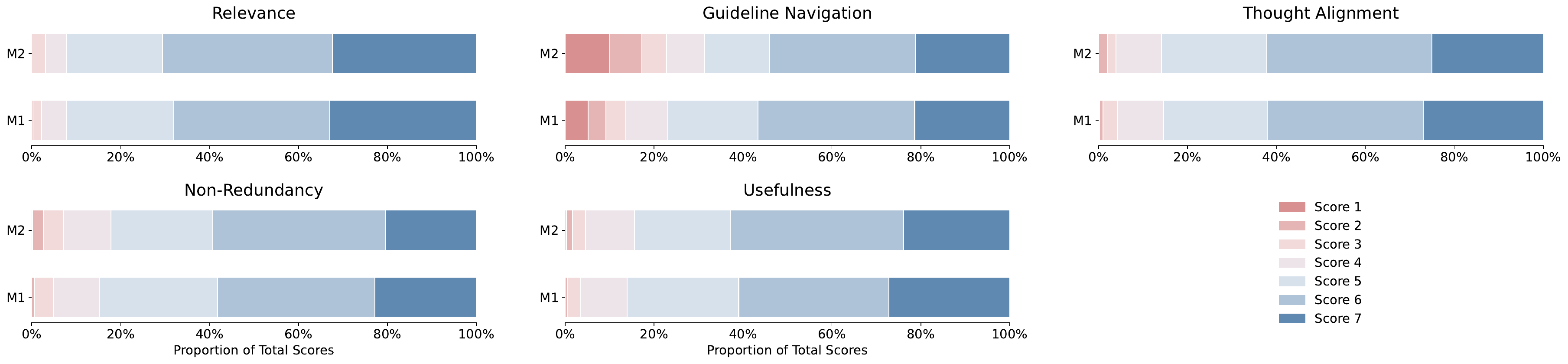}}%
        \begin{picture}(\wd\imagebox,\ht\imagebox)
            \put(0,0){\usebox{\imagebox}}
            \put(5,\ht\imagebox){\textbf{(a)}}
        \end{picture}
        \\[1em]
        \sbox{\imagebox}{\includegraphics[width=\linewidth]{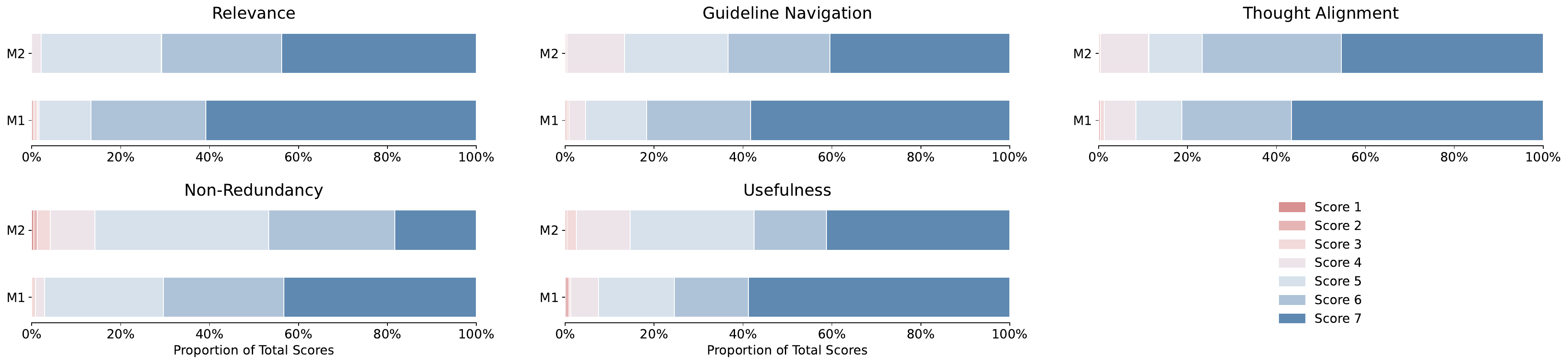}}%
        \begin{picture}(\wd\imagebox,\ht\imagebox)
            \put(0,0){\usebox{\imagebox}}
            \put(5,\ht\imagebox){\textbf{(b)}}
        \end{picture}
    \end{tabular}
    \caption{\textbf{Stacked bar plots of (a) the human evaluation by six clinicians and (b) the automated evaluation by Gemini on 80 dialogue transcripts.} M1 refers to the multi-stage reasoning method and M2 refers to the zero-shot baseline. Each method is rated on a Likert-scale from 1 – Strongly Disagree to 7 – Strongly Agree.}
    \label{fig:combined_eval}
\end{figure*}




%

\section{Results}

We present our main findings below.

\begin{figure}[ht]
    \centering
    \includegraphics[width=\linewidth]{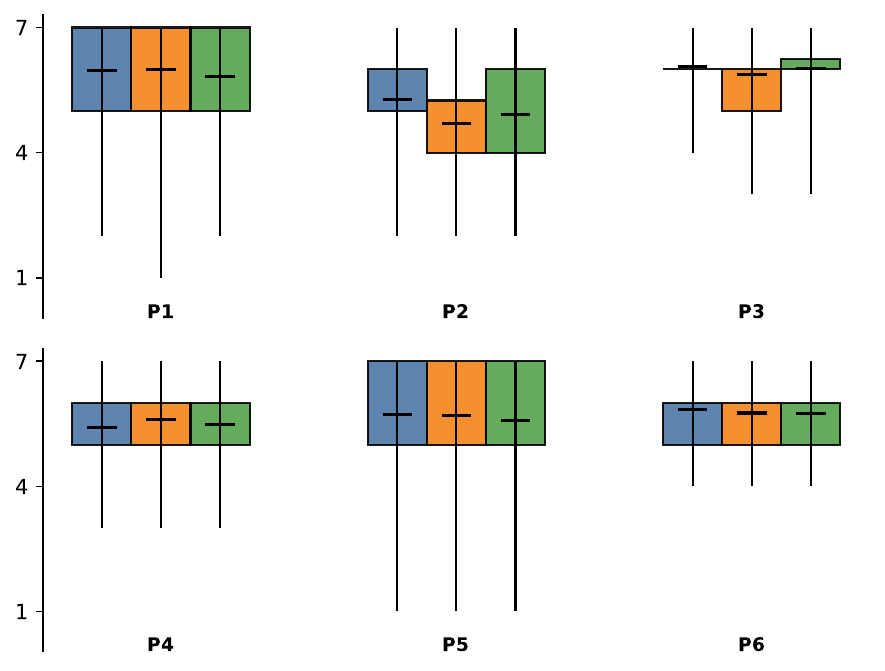}
    \caption{\textbf{Clinicians exhibit different rating styles.} On the X-axis, P1 to P6 means participated clinicians. The Y-axis, shows the range of Likert score for the evaluation result.
    Each group of box plots shows a clinician’s ratings of the multi-stage reasoning method across 40 annotated cases, evaluated under 30\%, 70\%, and 100\% dialogue context.}\label{fig:clinicians}
\end{figure}

\textbf{The AI System is perceived as a valuable tool by experienced clinicians.}
We find that experienced clinicians perceive our generated questions valuable in the context of routine patient visits. Across an evaluation of 80 cases, the proposed multi-stage reasoning framework received an average overall score of 5.63 on a 1-7 Likert scale, while the zero-shot baseline scored 5.54 (Figure~\ref{fig:combined_eval}a), which is a surprisingly strong result for an out-of-the-box LLM. On \textit{Usefulness} specifically, scores are 5.70 (proposed) and 5.65 (zero-shot). These scores are both well above the mid-point, which shows clinicians mostly agree the system is helpful and valuable (Figure~\ref{fig:clinicians}). We also measured the frequency with which clinicians commented \textit{``no question needed in the given context"} or \textit{``I don't find these questions useful"}. Across 1,440 data samples collected, fewer than 2\% received such comments, a very low rate of outright rejection. This quantitative reception is mirrored in the feedback, as one participating physician (P5) summarized their experience by noting that the AI's questions were \textit{``generally relevant and appropriate"}.
\begin{figure}[ht]
    \centering
    \includegraphics[width=\linewidth]{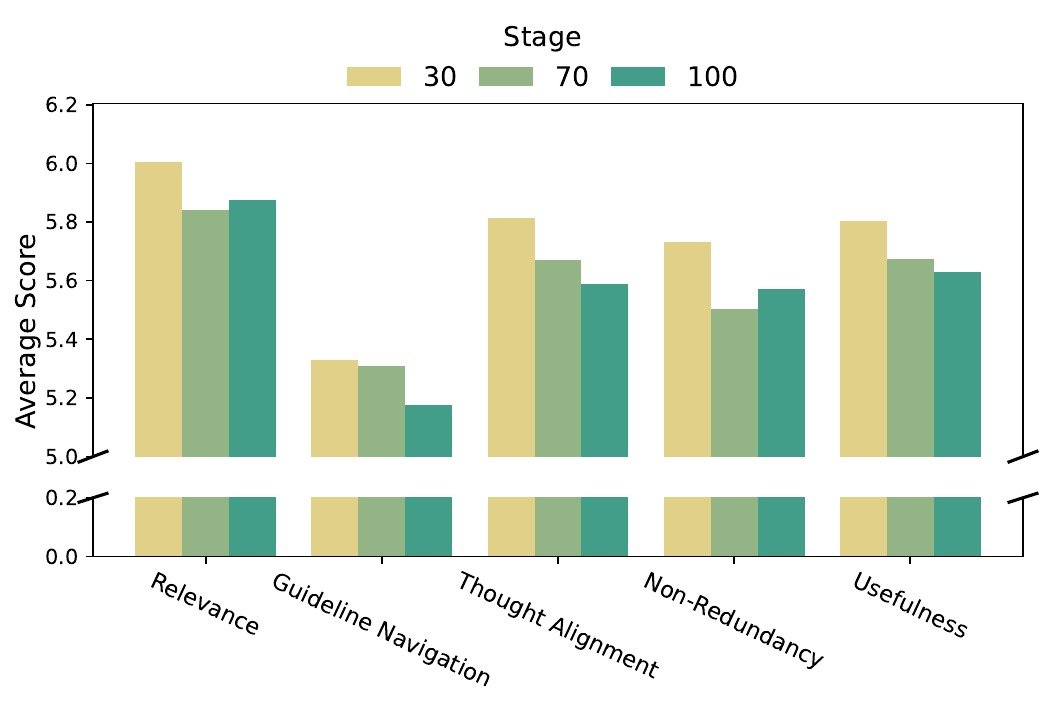}
    \caption{\textbf{Averaged scores of the proposed framework rated by PCPs at 30\%, 70\%, and 100\% dialogue context.} 
    }\label{fig:stage}
\end{figure}


\textbf{High-quality support is delivered consistently, even with partial context.} We analyzed human expert ratings of question quality at 30\%, 70\%, and 100\% of the dialogue (Figure~\ref{fig:stage}). Scores are remarkably stable: a one way ANOVA finds no significant differences across metrics or stages (all p-value $> 0.05$). This indicates our approach generates consistently high quality questions across varying context lengths. Notably, generations at 30\% context receive slightly higher scores across all five metrics, which aligns with pilot feedback that clinically decisive information often appears early and later turns can be redundant or noisy. This suggests the system surfaces nontrivial prompts even with limited context and is well suited for real time, ambient use throughout the visit rather than only post-hoc analysis tool.

%

\textbf{Multi-stage reasoning can enhance clinical safety and guideline alignment.} In high-stakes clinical environments, average performance is insufficient; reliability and safety are paramount. Our qualitative feedback underscored this point, as clinicians were highly sensitive to irrelevant or inaccurate suggestions. One physician (P5) noted that such prompts can ``take up valuable time in a visit," while another (P1) found unclear recommendations to be a significant barrier. These ``hallucinated" or poorly supported suggestions pose a direct risk to clinical utility and user trust. 
The zero-shot baseline produced low-quality or unsupported guideline citations (scores of 1-2 on Guideline Navigation) in 17.22\% of its generations versus 9.17\% for the proposed reasoning framework. Beyond this improvement, our method also yields small but consistent gains on four of five dimensions: Guideline Navigation (+6.72\%), Non-Redundancy (+1.51\%), Usefulness (+0.98\%), and Thought Alignment (+0.32\%). Relevance is essentially unchanged (–0.23\%), as shown in Figure~\ref{fig:combined_eval}a. Based on these findings, we conclude the multi-stage reasoning framework, while more complex than a zero-shot approach, is a valuable and preferable strategy for improving the reliability and safety in clinically-oriented tasks.

\begin{figure}[ht]
    \centering
    \includegraphics[width=\linewidth]{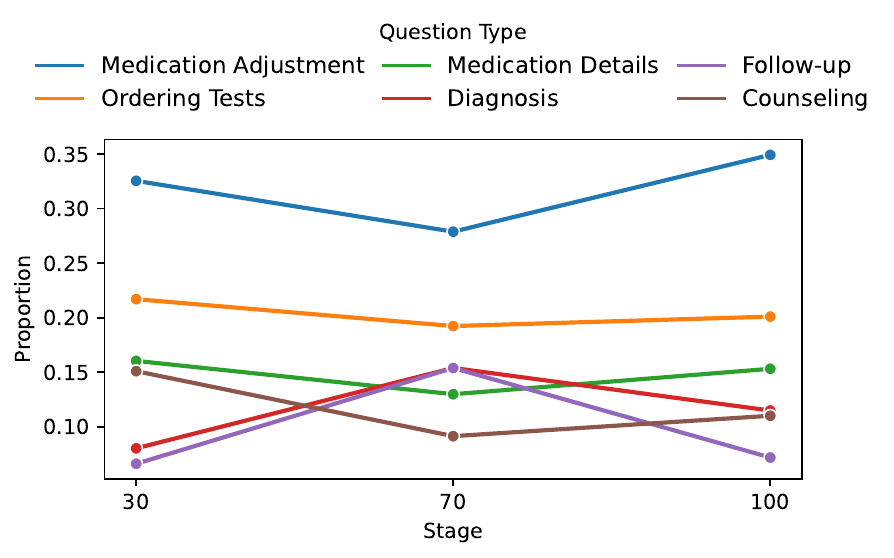}
    \caption{\textbf{Proportional trend of question types preferred by PCPs at 30\%, 70\%, and 100\% dialogue context.} Only questions generated by the proposed framework are included. The proportions may be biased by an uneven initial distribution of question types.
    }\label{fig:type}
\end{figure}

\vspace*{-2pt}
\textbf{Clinicians preferences for question types evolve during the encounter.} As mentioned in Section~\ref{sec:method:proposed}, the questions generated through our multi-stage reasoning framework fall into six distinct categories and offer diverse types of support for physicians. We compute each proportion as the number of times clinicians marked a question of that type as the ``best question" divided by the total number of valid clinician annotations at that context level. In Figure~\ref{fig:type}, the empirical pattern can be summarized in two points: (1) Across contexts clinicians show a persistent preference for questions that directly support management decisions, primarily medication adjustment and ordering tests. These categories dominate the distribution in both sparse and complete contexts (medication adjustment consistently above 25\%).
(2) A stage-dependent shift centered at 70\% context. At this stage, the dialogue introduces richer details, which prompts the clinician to shift focus toward specific follow-up actions (from 6\% to 15\%) and to begin forming preliminary judgments and diagnostic reasoning (from 8\% to 15\%), as clinicians seek to resolve residual uncertainty. Once the full context is present, preference returns to management-focused questions, and follow-up queries decline.

\textbf{LLM-as-judge reveals useful signal but with clear limitations.} We want to study whether there is an alternative or less expensive solution to reliably replicate clinicians’ evaluative priorities and safety judgments when assessing guideline-oriented questions. 

We instruct Gemini-2.5 Pro to evaluate two sets of our generated questions, following the exact same setting. 
We collect the results in a similar stacked bar plots on the same 80 dialogue transcripts in Figure~\ref{fig:combined_eval}b. LLM-as-judge shows our method outperforms the zero-shot baseline on all five metrics: Relevance +5.17\%, Guideline Navigation +7.49\%, Thought Alignment +2.87\%, Non-Redundancy +11.93\%, and Usefulness +7.46\%. Overall, the LLM assigns higher average scores (6.28 and 5.88 for ours and zero-shot, respectively).

Human evaluators and the LLM-as-judge agree on the direction of effect: both mark the proposed framework as preferable to the zero-shot baseline. In that sense the automated judge may offer a usable \textit{relative} comparison signal. However, the automatic scores diverge substantially from clinician ratings when examined quantitatively. See details in Appendix~\ref{apd:medgemma} and ~\ref{apd:correlation}. We highlight the following findings:

\begin{itemize}
    \item[1)] \textit{Systematic optimism.} Despite directional agreement, the LLM consistently inflates the magnitude of improvement. Automatic scores are also noticeably higher than human evaluation across multiple dimensions. Such bias suggests that while the LLM is a promising rapid proxy for comparative evaluation, its absolute judgments should be interpreted with caution.
    \vspace*{-1.5pt}
    \item[2)] \textit{Safety alignment risk.} The LLM-as-judge does not reliably detect certain classes of evidence- or guideline-related errors (\textit{i.e.,} low-quality outputs identified as false positives) that human reviewers do.
    \vspace*{-1.5pt}
    \item[3)] \textit{Deployment plausibility.} These observations imply a mixed deployment outlook: LLM can scale well and relatively agree with clinicians on the direction of effect, which is valuable for rapid, large-scale iteration; but its optimistic bias and hallucinations reduce its suitability as a standalone assessor for clinical safety. Human experts therefore remain the gold standard in such evidence-based, context-aware evaluation task.
    \vspace*{-1.5pt}
\end{itemize}

\textbf{Recruiting multiple clinicians to evaluate each case is crucial because clinical experts vary despite receiving the same rater instructions and having similar practices.} All of our clinicians are primary care doctors or internists with substantial experience (average years in practice: 16.5, IQR: [13, 21]) and are currently practicing. Nevertheless, their rating styles differ markedly (Figure~\ref{fig:clinicians}): P1 and P5 show wide interquartile ranges (5 and 7) and often assign low ratings of 1 or 2, whereas P4 and P6 rarely go below 3 or 4 and concentrate their scores around the mean. These differences highlight that clinicians bring individual tendencies and backgrounds to the task, so relying on a single annotator risks bias; multiple independent evaluations are essential for robust and reliable assessment in human-in-the-loop studies.


%




\section{Limitations}

Several limitations should be considered before real-world deployment of our system.  

\textbf{Cost.} Expert evaluation is resource-intensive; our 90 hours of physician review cost more than \$10k, which is not sustainable at scale. While API calls are cheaper than in-person visits, multi-stage prompting significantly increases token usage compared to zero-shot baselines, creating trade-offs between quality and expense.  

\textbf{Latency.} Our framework adds noticeable processing time (around 60 seconds to generate 3 questions in the multi-agent framework) due to API calls. Although acceptable offline, real-world use would also require audio transcription and integration into busy workflows. Such delays could limit the practicality of real-time ambient support.

\textbf{Generalizability.} Our evaluation is limited to 80 primary-care cases assessed by PCPs, so the results may not generalize to other specialties or broader case mixes. Many specialties pose different EBM challenges, for example, dermatology and radiology rely heavily on visual inputs, where text-centric LLMs may not meet without multimodal integration.


\section{Discussion and Future Work}
We explore whether LLMs can act as ambient assistants that surface evidence-based prompts during primary care encounters. Our multi-stage reasoning framework produces useful questions that align with guideline navigation approved by clinicians, while also revealing patterns in question-type preferences and partial alignment with LLM-as-judge. Overall, proactively surfacing guideline-relevant questions may ease cognitive load and support evidence-based reasoning in time-constrained visits, though real-world use remains limited by cost, latency, and privacy, with human experts as the gold standard for safety.

Future work can explore (1) \textit{proactivity}, by learning when to surface questions, what types to ask, and when to abstain in order to minimize burden, and (2) \textit{question answering}, by grounding responses in trusted guideline sources and delivering concise, actionable recommendations with provenance for safe clinical use.

\bibliography{jmlr-sample}

@article{sellergren2025medgemma,
  title={Medgemma technical report},
  author={Sellergren, Andrew and Kazemzadeh, Sahar and Jaroensri, Tiam and Kiraly, Atilla and Traverse, Madeleine and Kohlberger, Timo and Xu, Shawn and Jamil, Fayaz and Hughes, C{\'\i}an and Lau, Charles and others},
  journal={arXiv preprint arXiv:2507.05201},
  year={2025}
}

@article{neprash2021measuring,
  title={Measuring primary care exam length using electronic health record data},
  author={Neprash, Hannah T and Everhart, Alexander and McAlpine, Donna and Smith, Laura Barrie and Sheridan, Bethany and Cross, Dori A},
  journal={Medical care},
  volume={59},
  number={1},
  pages={62--66},
  year={2021},
  publisher={LWW}
}

@article{pondicherry2023designing,
  title={Designing clinical guidelines that improve access and satisfaction in the emergency department},
  author={Pondicherry, Neha and Schwartz, Hope and Stark, Nicholas and Dhanoa, Jaskirat and Emanuels, David and Singh, Malini and Peabody, Christopher R},
  journal={JACEP Open},
  volume={4},
  number={2},
  pages={e12919},
  year={2023},
  publisher={Elsevier}
}

@article{sutton2020overview,
  title={An overview of clinical decision support systems: benefits, risks, and strategies for success},
  author={Sutton, Reed T and Pincock, David and Baumgart, Daniel C and Sadowski, Daniel C and Fedorak, Richard N and Kroeker, Karen I},
  journal={NPJ digital medicine},
  volume={3},
  number={1},
  pages={17},
  year={2020},
  publisher={Nature Publishing Group UK London}
}

@inproceedings{mitchell2020development,
  title={The Development of a Point of Care Clinical Guidelines Mobile Application Following a User-Centred Design Approach},
  author={Mitchell, James and de Quincey, Ed and Pantin, Charles and Mustfa, Naveed},
  booktitle={International Conference on Human-Computer Interaction},
  pages={294--313},
  year={2020},
  organization={Springer}
}

@article{bates2003ten,
  title={Ten commandments for effective clinical decision support: making the practice of evidence-based medicine a reality},
  author={Bates, David W and Kuperman, Gilad J and Wang, Samuel and Gandhi, Tejal and Kittler, Anne and Volk, Lynn and Spurr, Cynthia and Khorasani, Ramin and Tanasijevic, Milenko and Middleton, Blackford},
  journal={Journal of the American Medical Informatics Association},
  volume={10},
  number={6},
  pages={523--530},
  year={2003},
  publisher={BMJ Group BMA House, Tavistock Square, London, WC1H 9JR}
}

@article{hager2024evaluation,
  title={Evaluation and mitigation of the limitations of large language models in clinical decision-making},
  author={Hager, Paul and Jungmann, Friederike and Holland, Robbie and Bhagat, Kunal and Hubrecht, Inga and Knauer, Manuel and Vielhauer, Jakob and Makowski, Marcus and Braren, Rickmer and Kaissis, Georgios and others},
  journal={Nature medicine},
  volume={30},
  number={9},
  pages={2613--2622},
  year={2024},
  publisher={Nature Publishing Group US New York}
}

@article{saab2024capabilities,
  title={Capabilities of gemini models in medicine},
  author={Saab, Khaled and Tu, Tao and Weng, Wei-Hung and Tanno, Ryutaro and Stutz, David and Wulczyn, Ellery and Zhang, Fan and Strother, Tim and Park, Chunjong and Vedadi, Elahe and others},
  journal={arXiv preprint arXiv:2404.18416},
  year={2024}
}

@article{vedadi2025towards,
  title={Towards physician-centered oversight of conversational diagnostic AI},
  author={Vedadi, Elahe and Barrett, David and Harris, Natalie and Wulczyn, Ellery and Reddy, Shashir and Ruparel, Roma and Schaekermann, Mike and Strother, Tim and Tanno, Ryutaro and Sharma, Yash and others},
  journal={arXiv preprint arXiv:2507.15743},
  year={2025}
}

@article{lange2024lost,
  title={Lost in translation: Unveiling medical students' untold errors of medical history documentation},
  author={Lange, Silvan and Kr{\"u}ger, Nils and Warm, Maximilian and Buechel, Johanna and Genzel-Borovicz{\'e}ny, Orsolya and Fischer, Martin R and Dimitriadis, Konstantinos},
  journal={The clinical teacher},
  volume={21},
  number={4},
  pages={e13749},
  year={2024},
  publisher={Wiley Online Library}
}

@article{saab2025advancing,
  title={Advancing Conversational Diagnostic AI with Multimodal Reasoning},
  author={Saab, Khaled and Freyberg, Jan and Park, Chunjong and Strother, Tim and Cheng, Yong and Weng, Wei-Hung and Barrett, David GT and Stutz, David and Tomasev, Nenad and Palepu, Anil and others},
  journal={arXiv preprint arXiv:2505.04653},
  year={2025}
}

@article{panteli2019clinical,
  title={Clinical practice guidelines as a quality strategy},
  author={Panteli, Dimitra and Legido-Quigley, Helena and Reichebner, Christoph and Ollenschl{\"a}ger, G{\"u}nter and Sch{\"a}fer, Corinna and Busse, Reinhard},
  journal={Improving healthcare quality in Europe},
  pages={233},
  year={2019},
  publisher={World Health Organization, Copenhagen, Denmark}
}

@article{comanici2025gemini,
  title={Gemini 2.5: Pushing the frontier with advanced reasoning, multimodality, long context, and next generation agentic capabilities},
  author={Comanici, Gheorghe and Bieber, Eric and Schaekermann, Mike and Pasupat, Ice and Sachdeva, Noveen and Dhillon, Inderjit and Blistein, Marcel and Ram, Ori and Zhang, Dan and Rosen, Evan and others},
  journal={arXiv preprint arXiv:2507.06261},
  year={2025}
}

@article{podder2021soap,
  title={Soap notes.[updated 2021 sep 2]},
  author={Podder, V and Lew, V and Ghassemzadeh, S},
  journal={StatPearls [Internet]. StatPearls Publishing. Available from: https://www. ncbi. nlm. nih. gov/books/NBK482263},
  year={2021}
}

@article{dolin2001hl7,
  title={The HL7 clinical document architecture},
  author={Dolin, Robert H and Alschuler, Liora and Beebe, Calvin and Biron, Paul V and Boyer, Sandra Lee and Essin, Daniel and Kimber, Elliot and Lincoln, Tom and Mattison, John E},
  journal={Journal of the American Medical Informatics Association},
  volume={8},
  number={6},
  pages={552--569},
  year={2001},
  publisher={BMJ Group BMA House, Tavistock Square, London, WC1H 9JR}
}

@inproceedings{liu2025proactive,
  title={Proactive conversational agents with inner thoughts},
  author={Liu, Xingyu Bruce and Fang, Shitao and Shi, Weiyan and Wu, Chien-Sheng and Igarashi, Takeo and Chen, Xiang'Anthony'},
  booktitle={Proceedings of the 2025 CHI Conference on Human Factors in Computing Systems},
  pages={1--19},
  year={2025}
}

@article{ayaz2021fast,
  title={The Fast Health Interoperability Resources (FHIR) standard: systematic literature review of implementations, applications, challenges and opportunities},
  author={Ayaz, Muhammad and Pasha, Muhammad F and Alzahrani, Mohammed Y and Budiarto, Rahmat and Stiawan, Deris},
  journal={JMIR medical informatics},
  volume={9},
  number={7},
  pages={e21929},
  year={2021},
  publisher={JMIR Publications Toronto, Canada}
}

@article{ferber2024gpt,
  title={GPT-4 for information retrieval and comparison of medical oncology guidelines},
  author={Ferber, Dyke and Wiest, Isabella C and W{\"o}lflein, Georg and Ebert, Matthias P and Beutel, Gernot and Eckardt, Jan-Niklas and Truhn, Daniel and Springfeld, Christoph and J{\"a}ger, Dirk and Kather, Jakob Nikolas},
  journal={Nejm Ai},
  volume={1},
  number={6},
  pages={AIcs2300235},
  year={2024},
  publisher={Massachusetts Medical Society}
}

@inproceedings{oniani2024enhancing,
  title={Enhancing large language models for clinical decision support by incorporating clinical practice guidelines},
  author={Oniani, David and Wu, Xizhi and Visweswaran, Shyam and Kapoor, Sumit and Kooragayalu, Shravan and Polanska, Katelyn and Wang, Yanshan},
  booktitle={2024 IEEE 12th International Conference on Healthcare Informatics (ICHI)},
  pages={694--702},
  year={2024},
  organization={IEEE}
}

@article{fast2024autonomous,
  title={Autonomous medical evaluation for guideline adherence of large language models},
  author={Fast, Dennis and Adams, Lisa C and Busch, Felix and Fallon, Conor and Huppertz, Marc and Siepmann, Robert and Prucker, Philipp and Bayerl, Nadine and Truhn, Daniel and Makowski, Marcus and others},
  journal={NPJ Digital Medicine},
  volume={7},
  number={1},
  pages={358},
  year={2024},
  publisher={Nature Publishing Group UK London}
}

@article{lichtner2023automated,
  title={Automated monitoring of adherence to evidenced-based clinical guideline recommendations: design and implementation study},
  author={Lichtner, Gregor and Spies, Claudia and Jurth, Carlo and Bienert, Thomas and Mueller, Anika and Kumpf, Oliver and Piechotta, Vanessa and Skoetz, Nicole and Nothacker, Monika and Boeker, Martin and others},
  journal={Journal of Medical Internet Research},
  volume={25},
  pages={e41177},
  year={2023},
  publisher={JMIR Publications Toronto, Canada}
}

@article{sagheb2022artificial,
  title={Artificial intelligence assesses clinicians’ adherence to asthma guidelines using electronic health records},
  author={Sagheb, Elham and Wi, Chung-Il and Yoon, Jungwon and Seol, Hee Yun and Shrestha, Pragya and Ryu, Euijung and Park, Miguel and Yawn, Barbara and Liu, Hongfang and Homme, Jason and others},
  journal={The Journal of Allergy and Clinical Immunology: In Practice},
  volume={10},
  number={4},
  pages={1047--1056},
  year={2022},
  publisher={Elsevier}
}

@article{Yarnall2003,
  author = {Yarnall, K. S. H. and {\O}stbye, T. and Krause, K. M. and Pollak, K. I. and Gradison, M. and Michener, J. L.},
  title = {Primary care: Is there enough time for prevention?},
  journal = {Am. J. Public Health},
  year = {2003},
  volume = {93},
  number = {4},
  pages = {635--641},
  doi = {10.2105/AJPH.93.4.635}
}

@article{Ostbye2005,
  author = {{\O}stbye, T. and Yarnall, K. S. H. and Krause, K. M. and Pollak, K. I. and Gradison, M. and Michener, J. L.},
  title = {Is there time for management of patients with chronic diseases in primary care?},
  journal = {Ann. Fam. Med.},
  year = {2005},
  volume = {3},
  number = {3},
  pages = {209--214},
  doi = {10.1370/afm.285}
}

@article{Skandari2023,
  author = {Porter, J. and Boyd, C. and Skandari, M. R. and Laiteerapong, N.},
  title = {Revisiting the time needed to provide adult primary care},
  journal = {J. Gen. Intern. Med.},
  year = {2023},
  volume = {38},
  number = {1},
  pages = {147--155},
  doi = {10.1007/s11606-022-07707-x}
}

@article{Petterson2012,
  author = {Petterson, S. M. and Liaw, W. R. and Phillips, R. L. and Rabin, D. L. and Meyers, D. S. and Bazemore, A. W.},
  title = {Projecting US primary care physician workforce needs: 2010-2025},
  journal = {Ann. Fam. Med.},
  year = {2012},
  volume = {10},
  number = {6},
  pages = {503--509},
  doi = {10.1370/afm.1461}
}

@article{Zwolsman2012,
  author = {Zwolsman, S. and te Pas, E. and Hooft, L. and Wieringa-de Waard, M. and van Dijk, N.},
  title = {Barriers to GPs' use of evidence-based medicine: a systematic review},
  journal = {Br. J. Gen. Pract.},
  year = {2012},
  volume = {62},
  number = {600},
  pages = {e511--e521},
  doi = {10.3399/bjgp12X652382}
}

@article{DelFiol2014,
  author = {Del Fiol, G. and Workman, T. E. and Gorman, P. N.},
  title = {Clinical questions raised by clinicians at the point of care: a systematic review},
  journal = {JAMA Intern. Med.},
  year = {2014},
  volume = {174},
  number = {5},
  pages = {710--718},
  doi = {10.1001/jamainternmed.2014.368}
}

@article{McGinn2013,
  author = {McGinn, T. G. and McCullagh, L. and Kannry, J. and Knaus, M. and Sofianou, A. and Wisnivesky, J. P. and Mann, D. M.},
  title = {Efficacy of an evidence-based clinical decision support in primary care practices: a randomized clinical trial},
  journal = {JAMA Intern. Med.},
  year = {2013},
  volume = {173},
  number = {17},
  pages = {1584--1591},
  doi = {10.1001/jamainternmed.2013.8980}
}

@article{Garg2005,
  author = {Garg, A. X. and Adhikari, N. K. and McDonald, H. and Rosas-Arellano, M. P. and Devereaux, P. J. and Beyene, J. and Sam, J. and Haynes, R. B.},
  title = {Effects of computerized clinical decision support systems on practitioner performance and patient outcomes: a systematic review},
  journal = {JAMA},
  year = {2005},
  volume = {293},
  number = {10},
  pages = {1223--1238},
  doi = {10.1001/jama.293.10.1223}
}

@article{Kawamoto2005,
  author = {Kawamoto, K. and Houlihan, C. A. and Balas, E. A. and Lobach, D. F.},
  title = {Improving clinical practice using clinical decision support systems: a systematic review of trials to identify features critical to success},
  journal = {BMJ},
  year = {2005},
  volume = {330},
  number = {7494},
  pages = {765},
  doi = {10.1136/bmj.38398.500764.8F}
}

@article{Vasey2021,
  author = {Vasey, B. and Ursprung, S. and Beddoe, B. and Taylor, P. and Marlow, N. and Bilbro, G. and Watkinson, P. and McCulloch, P.},
  title = {Association of clinician diagnostic performance with machine learning-based decision support systems: a systematic review},
  journal = {JAMA Netw. Open},
  year = {2021},
  volume = {4},
  number = {3},
  pages = {e211474},
  doi = {10.1001/jamanetworkopen.2021.1474}
}

@article{FraileNavarro2025,
  author = {Fraile Navarro, D. and Coiera, E. and Hambly, T. W. and Triplett, Z. and Asif, N. and Susanto, A. and Chowdhury, A. and Azcoaga Lorenzo, A. and Dras, M. and Berkovsky, S.},
  title = {Expert evaluation of large language models for clinical dialogue summarization},
  journal = {Sci. Rep.},
  year = {2025},
  volume = {15},
  pages = {1195},
  doi = {10.1038/s41598-024-84850-x}
}

@article{VanVeen2024,
  author = {Van Veen, D. and Van Uden, C. and Blankemeier, L. and Delbrouck, J.-B. and Aali, A. and Bluethgen, C. and Pareek, A. and Polacin, M. and Reis, E. P. and Seehofnerov\'{a}, A. and Rohatgi, N. and Hosamani, P. and Collins, W. and Ahuja, N. and Langlotz, C. P. and Hom, J. and Gatidis, S. and Pauly, J. and Chaudhari, A. S.},
  title = {Clinical text summarization: adapting large language models can outperform human experts},
  journal = {Nat. Med.},
  year = {2024},
  volume = {30},
  number = {4},
  pages = {1134--1142},
  doi = {10.1038/s41591-024-02855-5}
}

@article{Singhal2025,
  author = {Singhal, K. and Tu, T. and Gottweis, J. and Sayres, R. and Wulczyn, E. and Natarajan, V. and Azizi, S. and Karthikesalingam, A. and Liu, Y. and others},
  title = {Toward expert-level medical question answering with large language models},
  journal = {Nat. Med.},
  year = {2025},
  volume = {31},
  number = {5},
  pages = {943--950},
  doi = {10.1038/s41591-024-03423-7}
}

@article{Ayers2023,
  author = {Ayers, J. W. and Poliak, A. and Dredze, M. and Leas, E. C. and Zhu, Z. and Kelley, J. B. and Faix, D. J. and Goodman, A. M. and Longhurst, C. A. and Hogarth, M. and Smith, D. M.},
  title = {Comparing physician and artificial intelligence chatbot responses to patient questions posted to a public social media forum},
  journal = {JAMA Intern. Med.},
  year = {2023},
  volume = {183},
  number = {6},
  pages = {589--596},
  doi = {10.1001/jamainternmed.2023.1838}
}

@article{Shi2024,
  author = {Shi, Y. and Xu, S. and Yang, T. and Liu, Z. and Liu, T. and Li, Q. and Li, X. and Liu, N.},
  title = {{MKRAG}: Medical knowledge retrieval augmented generation for medical question answering},
  journal = {arXiv preprint arXiv:2309.16035},
  year = {2024}
}

@article{Li2024,
  author = {Li, S. S. and Balachandran, V. and Feng, S. and Ilgen, J. and Pierson, E. and Koh, P. W. and Tsvetkov, Y.},
  title = {{MediQ}: Question-asking {LLMs} for adaptive and reliable clinical reasoning},
  journal = {arXiv preprint arXiv:2406.00922},
  year = {2024}
}

@article{Tu2025,
  author = {Tu, T. and Schaekermann, M. and Palepu, A. and Saab, K. and Freyberg, J. and others},
  title = {Towards conversational diagnostic artificial intelligence},
  journal = {Nature},
  year = {2025},
  volume = {642},
  pages = {442--450},
  doi = {10.1038/s41586-025-08866-7}
}

@article{Goh2025a,
  author = {Goh, E. and Gallo, R. J. and Strong, E. and Weng, Y. and Kerman, H. and Chen, J. H. and others},
  title = {{GPT-4} assistance for improvement of physician performance on patient care tasks: a randomized controlled trial},
  journal = {Nat. Med.},
  year = {2025},
  volume = {31},
  number = {4},
  pages = {1233--1238},
  doi = {10.1038/s41591-024-03456-y}
}

@article{Goh2025b,
  author = {Goh, E. and Bunning, B. and Khoong, E. C. and Gallo, R. J. and Milstein, A. and Chen, J. H. and others},
  title = {Physician clinical decision modification and bias assessment in a randomized controlled trial of {AI} assistance},
  journal = {Commun. Med.},
  year = {2025},
  volume = {5},
  pages = {59},
  doi = {10.1038/s43856-025-00781-2}
}

@article{Hager2024,
  author = {Hager, P. and Jungmann, F. and Holland, R. and Bhagat, K. and Hubrecht, I. and Kaissis, G. and Rueckert, D. and others},
  title = {Evaluation and mitigation of the limitations of large language models in clinical decision-making},
  journal = {Nat. Med.},
  year = {2024},
  volume = {30},
  pages = {2613--2622},
  doi = {10.1038/s41591-024-03097-1}
}

@article{Wang2025,
  author = {Wang, Ziyu and Li, Hao and Huang, Di and Kim, Hye-Sung and Shin, Chae-Won and Rahmani, Amir M.},
  title = {HealthQ: Unveiling Questioning Capabilities of LLM Chains in Healthcare Conversations},
  journal = {Smart Health},
  year = {2025},
  volume = {36},
  pages = {100570},
  doi = {10.1016/j.smhl.2025.100570}
}

@article{xie2025kerap,
  title={KERAP: A Knowledge-Enhanced Reasoning Approach for Accurate Zero-shot Diagnosis Prediction Using Multi-agent LLMs},
  author={Xie, Yuzhang and Cui, Hejie and Zhang, Ziyang and Lu, Jiaying and Shu, Kai and Nahab, Fadi and Hu, Xiao and Yang, Carl},
  journal={arXiv preprint arXiv:2507.02773},
  year={2025}
}

@inproceedings{Gatto2025,
  author = {Gatto, Joseph and Seegmiller, Parker and Burdick, Timothy and Khayal, Inas S. and DeLozier, Sarah and Preum, Sarah M.},
  title = {Follow-up Question Generation For Enhanced Patient-Provider Conversations},
  booktitle = {Proc. of the 63rd Annual Meeting of the Association for Computational Linguistics (ACL)},
  year = {2025},
  pages = {25222–25240},
  url = {https://aclanthology.org/2025.acl-long.1226.pdf}
}

@article{ashman2023characteristics,
  title={Characteristics of office-based physician visits by age, 2019},
  author={Ashman, Jill J and Santo, Loredana and Okeyode, Titilayo},
  year={2023}
}

\newpage

\appendix

\newpage

\section{Prompt Template}\label{apd:first}

We provide our prompt templates described in Section~\ref{sec:method}. Table~\ref{tb:agent1}, \ref{tb:agent2}, \ref{tb:agent3} are the prompts for three agents in the multi-stage reasoning framework. Table~\ref{tb:zs} is the prompt for the zero-shot setting.

\begin{table*}
  \centering
  \begin{simplebox}
    \textbf{Prompt:} 
    You are an expert medical assistant specializing in clinical documentation. Your task is to analyze the provided doctor-patient dialogue and transform it into a structured, clinically relevant summary in JSON format. Base your summary ONLY on the information present in the dialogue. Do not infer conditions or details that are not explicitly stated or directly implied. Use concise, clear, professional language. Specific numerical information in the dialogue must be fully recorded, such as medication dosage, BMI, age, etc.\\[5pt]
Analyze the dialogue provided below in the \(<\)dialogue\(>\) section and the EHR from the \(<\)questionnaire\(>\) section. Populate the following JSON structure based on these field definitions:\\[5pt]
\hangitem{- chiefComplaint: The primary reason for the visit, as stated by the patient. Keep it brief.}
\vspace{5pt}
\hangitem{- historyOfPresentIllness: A detailed narrative of the chief complaint. Include descriptions of symptoms (quality, timing, severity), pertinent negatives (symbols the patient denies having), and any other related complaints explored during the visit.}
\vspace{5pt}
\hangitem{- medicationHistory: List all medications mentioned, including the ones in the conversation and patient’s past history. Note their reported effectiveness if mentioned.}
\vspace{5pt}
\hangitem{- personalHistory: List known chronic conditions (Past Medical History) and key lifestyle factors (Social History) like smoking status.}
\vspace{5pt}
\hangitem{- objectiveFindings: List all objective, observable findings from the doctor's physical exam and any mentioned vitals.}
\vspace{5pt}
\hangitem{- doctorsAssessment: The final diagnoses or clinical conclusions reached by the doctor.}
\vspace{5pt}
\hangitem{- plan: The course of action decided upon. Include any new prescriptions, tests ordered, and patient education provided in the dialogue.}
\vspace{5pt}
\(<\)DIALOGUE\(>\){{content}}\(<\)/DIALOGUE\(>\)\\[5pt]
\(<\)QUESTIONNAIRE\(>\){{questionnaire}}\(<\)/QUESTIONNAIRE\(>\)

  \end{simplebox}
  
  \caption{Prompt for the summarization agent.}
  \label{tb:agent1}
\end{table*}

\begin{table*}
  \centering
  \begin{simplebox}
    \textbf{Prompt:} 
    You are an experienced primary care physician shadowing a peer during their patient visit. Your primary role is to be helpful and supportive, using your clinical experience as a `second set of eyes' on the case. The purpose is to generate search/reference-oriented questions that reflect the problems physicians face when consulting medical guidelines.
    
\vspace{5pt}

Your task is to analyze the provided clinical summary and generate exactly 10 distinct questions. The questions must be grounded in the details of the provided context. The questions should cover the following categories:\\[5pt]
\hspace*{14pt}* Medication Adjustment\\
\hspace*{14pt}* Ordering Tests\\
\hspace*{14pt}* Medication Details (e.g., dosage, use, adverse effects)\\
\hspace*{14pt}* Diagnosis (e.g., differential diagnosis)\\
\hspace*{14pt}* Follow-up\\
\hspace*{14pt}* Counseling\\[5pt]
Instructions:
\vspace{5pt}
\hangitem{- Purpose: The goal is to create questions that are search/reference-oriented. They should be the kind of questions a doctor would use to query a clinical evidence database or guideline repository.}
\vspace{5pt}
\hangitem{- Structure and Style: Each of the 10 questions you generate must be a detailed, single-paragraph clinical vignette. Frame them from the professional perspective of a physician and use standard medical terminology.}
\vspace{5pt}
\hangitem{- Embed Context: Each question must contain all the necessary context for it to be a standalone query. An agent answering the question should not need the original summary. Follow this structure within the question:}
\vspace{5pt}
\hspace*{28pt}1.  Patient Profile: (e.g., ``A 62-year-old male...")\\
\hspace*{28pt}2.  Relevant History \& Diagnoses: (e.g., ``...with hypertension and a history of gout...")\\
\hspace*{28pt}3.  Current Clinical Status: (e.g., ``...on Lisinopril 20mg daily...")\\
\hspace*{28pt}4.  The Core Guideline Question: (e.g., ``according to the 2017 ACC/AHA guidelines")
\vspace{5pt}
\hangitem{- Formatting: Do not include the question category (e.g., ``Medication Adjustment", ``Ordering Tests") in the question text itself. Generate only a numbered list of questions. Do not provide answers. Each question must be a single, focused query. Do not chain multiple questions together using conjunctions like `and' or `or'. Ensure each string in the list poses one, and only one, question.}
\vspace{5pt}
\hangitem{- Quantity: Ensure there are exactly 10 questions generated from the single summary provided.}
\vspace{5pt}
\hangitem{- Tone and Persona: Frame each question as if you are a helpful, collaborative clinical partner. Your goal is to gently remind or prompt for deeper thinking, not to test the user. Use collaborative phrasing where appropriate. The tone should be supportive and aim to reduce, not increase, the user's cognitive load.}
\vspace{5pt}
EXAMPLE 1: A 6-year-old child has had persistent coryza, mild cough, and low-grade fever for 12 days. What, if any, diagnostic tests are recommended by guidelines at this point for an uncomplicated but prolonged URI in a child, before considering empiric antibiotics for suspected sinusitis?\\[5pt]
EXAMPLE 2: A 58-year-old female with Type 2 Diabetes, hypertension, and hyperlipidemia is currently on metformin 1000mg twice daily, lisinopril 10mg daily, and atorvastatin 40mg daily. Her latest A1c is 8.5\%, and her LDL-C is 110 mg/dL. As per the latest ADA guidelines, what is the recommended second-line agent to add for glycemic control, considering her cardiovascular risk factors?\\[5pt]
IMPORTANT NOTE: The two examples above are for style and format reference only. The clinical details within them are not related to the actual case provided below. Your task is to generate 10 new questions based only on the following summary.\\[5pt]
\(<\)SUMMARY\(>\){summary}\(<\)/SUMMARY\(>\)\\[5pt]

  \end{simplebox}
  
  \caption{Prompt for the question raising agent.}
  \label{tb:agent2}
\end{table*}

\begin{table*}
  \centering
  \begin{simplebox}
    \textbf{Prompt:} 
You are a meticulous AI Quality Assurance specialist. Your task is to evaluate a batch of candidate questions against a provided clinical dialogue summary. For each question, you will first reason about its pros and cons, and then assign a score for each of the seven distinct criteria, finally calculating an average score.\\[5pt]
Instructions:\\[5pt]
\hspace*{14pt}1. Process Sequentially: Evaluate each of the questions in the `candidateQuestions' list independently.\\
\hspace*{14pt}2. Reason Before Rating: For each question, you must first perform the positive and negative reasoning steps before assigning scores.\\
\hspace*{14pt}3. Adhere to the Scale: You MUST score each criterion on a scale of 1.0 to 5.0, using increments of 0.5.\\
\hspace*{14pt}4. Strictly Follow the Output Format: Your final output must be a single, valid JSON object.\\[5pt]
\(<\)SUMMARY\(>\){summary}\(<\)/SUMMARY\(>\)\\[5pt]
\(<\)QUESTIONS\(>\){generated\_questions}\(<\)/QUESTIONS\(>\)\\[5pt]
For EACH question in the `candidateQuestions' list, perform the following steps in order:\\[5pt]
1. Positive Assessment (Pros):
   First, reason about why the question is good and useful for a doctor.
   From the seven criteria below, select the top two that best argue FOR the question's quality and briefly state why.\\[5pt]
2. Negative Assessment (Cons):
   Second, reason about why the question is not good enough or could be improved.
   From the seven criteria below, select the top two that best argue AGAINST the question's quality and briefly state why.\\[5pt]
3. Detailed Scoring:
   Third, based on your balanced assessment from the steps above, assign a score from 1.0 to 5.0 for ALL seven of the following criteria.
\vspace{5pt}
\hangitem{* relevance: How relevant and important is the question to the patient's chief complaint and core clinical problems identified in the summary?}
\vspace{5pt}
\hangitem{* expectedImpact: How likely is the question to save the doctor time, uncover critical missing information, or directly improve the quality of the clinical decision?}
\vspace{5pt}
\hangitem{* originality: Is the question redundant? Does it ask something already answered or made obvious by the summary?.}
\vspace{5pt}
\hangitem{* factualAccuracy: Is the premise of the question factually consistent with the summary?}
\vspace{5pt}
\hangitem{* comprehensiveness: Is the question itself well-formed and complete enough to be answerable, or is it too vague?}
\vspace{5pt}
\hangitem{* clarityAndConciseness: Is the question clear, professional, and easy to understand?}
\vspace{5pt}
\hangitem{* collaborativeTone: How well is the question framed to be supportive, non-confrontational, and respectful of the physician's expertise? (A low score means the question sounds like a test or a command).}
\vspace{5pt}
4. Mean Score Calculation:
   Finally, calculate the `meanScore' by averaging the seven individual scores from Step 3. Round the result to two decimal places.\\[5pt]
5. Format the Output:
   Consolidate all the information for the question into a single JSON object as specified below and add it to the `evaluationResults' list. Repeat for all questions.\\[5pt]
OUTPUT FORMAT:\\[5pt]
Your entire output must be a single JSON object. Do not add any text before or after it.

  \end{simplebox}
  
  \caption{Prompt for the question evaluation agent.}
  \label{tb:agent3}
\end{table*}

\begin{table*}
  \centering
  \begin{simplebox}
    \textbf{Prompt:} 
    You are an experienced primary care physician shadowing a peer during their patient visit. Your primary role is to be helpful and supportive, using your clinical experience to provide useful and meaningful questions. Your questions should be search/reference-oriented questions that reflect the problems physicians face when consulting medical guidelines.\\[5pt]
Please analyze the provided clinical dialogue (given under \(<\)DIALOGUE\(>\) and \(<\)QUESTIONNAIRE\(>\) section) and generate exactly 3 distinct questions. The questions must be grounded in the details of the provided dialogue. Each question should be a standalone, context-rich clinical vignette. Each question must be a single, focused query. Do not chain multiple questions together using conjunctions like `and' or `or'. Ensure each string in the list poses only one question. Generate only a numbered list of questions. Do not provide answers.\\[5pt]
\(<\)DIALOGUE\(>\){{content}}\(<\)/DIALOGUE\(>\)\\[5pt]
\(<\)QUESTIONNAIRE\(>\){{questionnaire}}\(<\)/QUESTIONNAIRE\(>\)\\[7pt]
OUTPUT:\\[5pt]
Your entire output must be a single, valid JSON object with no additional text before or after it. The JSON object must contain a single key, ``questions", with a list of exactly three strings.
      \end{simplebox}
  
  \caption{Prompt for the zero-shot baseline setting.}
  \label{tb:zs}
\end{table*}

\section{Human Evaluation Instructions for Clinician Raters}\label{apd:instructions}

\textbf{Background.} This evaluation focuses on an Evidence-Based Medical Guideline Agent, an AI tool designed to raise useful, insightful, and reference-oriented questions that reflect the problems physicians face when consulting medical guidelines. The purpose of developing this agent is to assist primary care physicians during their outpatient visits. In these brief encounters, physicians are expected to make informed decisions and ensure their practice aligns with current medical guidelines.

\vspace*{6pt}
\noindent \textbf{Task.} The task is to evaluate the questions generated by different AI agents based on real-world physician-patient dialogues. Assuming you are the physician in the dialogue, you will need to evaluate two sets of questions generated by two AI agents for each dialogue. The same case will be evaluated under three dialogues of varying lengths. You will assess each set of questions against a rubric built on five key dimensions: Relevance, Guideline Navigation, Thought Alignment, Non-Redundancy, and Usefulness.

You should first begin by carefully reading the provided dialogue context and patient's questionnaire. Next, you will be provided with a set of three distinct questions. For this set of questions as a whole, you will use a 7-point Likert scale to score each metric. You are required to select the question that you believe is the most appropriate and beneficial within the given context. You can: (1) Choose a preferred question, (2) Select none, if you don’t need questions under such context, (3) Provide custom question / question type / any idea (optional).

\vspace*{6pt}

\section{Evaluation Interface for Human Evaluation}\label{apd:second}

We used a simple web based survey workflow. A central progress tracker listed the 80 cases, one row per case, with the assigned clinician, a completion checkbox, and a link to the case questionnaire (Figure~\ref{fig:eval-tracker}). Both annotators and the study team could see progress and access the corresponding questionnaire from this list.

Each case questionnaire had six pages. The six pages corresponded to two model variants at three dialogue context levels (Figure~\ref{fig:eval-questionnaire}). Raters saw the 30 percent context twice, the 70 percent context twice, and the 100 percent context twice. Each page began with the patient health record and the relevant portion of the patient clinician dialogue. When content had already been shown at an earlier page, a clear indicator stated that the patient record or the earlier dialogue had been presented, and a jump marker allowed the rater to scroll directly to the new content for that stage (Figure~\ref{fig:eval-questionnaire}).

Below the context, each page displayed three candidate questions labeled A, B, and C. Raters completed five seven point Likert items aligned with our metrics of Relevance, Guideline Navigation, Thought Alignment, Non Redundancy, and Usefulness. A final item asked the rater to select the single most useful question, with options A, B, C, no question needed, or Other. Choosing Other opened a free text field where raters could propose an alternative question or leave comments.

To mitigate order effects, approximately half of the cases presented the two model variants in reversed order at each context level. The interface kept wording and layout identical across pages and cases, and required a response for each item before advancing, which reduced missing data and kept the rating experience consistent across annotators.

\begin{figure*}[t]
\centering
\resizebox{\linewidth}{!}{%
\begin{tikzpicture}[x=1cm,y=1cm,font=\small]

\draw[rounded corners=2pt,fill=gray!10] (-0.2,0.2) rectangle (13.0,-6.2);

\begin{scope}[yshift=-0.6cm] 

\node[anchor=west,font=\bfseries] at (0,0) {Progress tracker for 80 cases};

\def\xA{0}    
\def\xB{2.4}  
\def\xC{6.2}  
\def\xD{8.2}  
\def\xE{12.8} 
\def\rowH{0.8}

\fill[gray!25,rounded corners=1pt] (\xA,-0.3) rectangle (\xE,-0.3-\rowH);
\draw (\xA,-0.3) rectangle (\xB,-0.3-\rowH);
\draw (\xB,-0.3) rectangle (\xC,-0.3-\rowH);
\draw (\xC,-0.3) rectangle (\xD,-0.3-\rowH);
\draw (\xD,-0.3) rectangle (\xE,-0.3-\rowH);
\node[anchor=west] at (0.1,-0.3-0.5*\rowH) {Case ID};
\node[anchor=west] at (\xB+0.1,-0.3-0.5*\rowH) {Clinician};
\node at ({(\xC+\xD)/2},-0.3-0.5*\rowH) {Done};
\node[anchor=west] at (\xD+0.1,-0.3-0.5*\rowH) {Questionnaire link};

\foreach \i/\case/\clin/\done in {1/001/Dr.\ A/\checkmark,2/002/Dr.\ B/,3/003/Dr.\ C/\checkmark,4/004/Dr.\ D/,5/005/Dr.\ E/} {
  \pgfmathsetmacro{\y}{-0.3-\i*\rowH}
  \fill[white] (\xA,\y) rectangle (\xE,\y-\rowH);
  \draw (\xA,\y) rectangle (\xB,\y-\rowH);
  \draw (\xB,\y) rectangle (\xC,\y-\rowH);
  \draw (\xC,\y) rectangle (\xD,\y-\rowH);
  \draw (\xD,\y) rectangle (\xE,\y-\rowH);

  \node[anchor=west] at (\xA+0.1,\y-0.5*\rowH) {Case \case};
  \node[anchor=west] at (\xB+0.1,\y-0.5*\rowH) {\clin};
  \draw (\xC+0.2,\y-0.2) rectangle (\xC+0.6,\y-0.6); 
  \ifx\done\empty\else \node at ({\xC+0.4},{\y-0.4}) {\done}; \fi
  \draw[rounded corners=2pt,fill=blue!10] (\xD+0.1,\y-0.2) rectangle (\xD+2.6,\y-0.6);
  \node[anchor=west,font=\scriptsize] at (\xD+0,\y-0.4) {Open questionnaire};
}

\end{scope}
\end{tikzpicture}
}
\caption{Progress list with one row per case, a completion checkbox, and a link to the questionnaire.}
\label{fig:eval-tracker}
\end{figure*}

\begin{figure*}[t]
\centering
\resizebox{\linewidth}{!}{%
\begin{tikzpicture}[x=1cm,y=1cm,font=\small]
\draw[rounded corners=3pt,fill=gray!10] (-0.2,0.2) rectangle (13.0,-10.3);

\draw[rounded corners=8pt,fill=green!15] (0.0,0.0) rectangle (1.4,-0.5);
\node at (0.7,-0.25) {\scriptsize Case 037};
\draw[rounded corners=8pt,fill=blue!15] (1.6,0.0) rectangle (3.6,-0.5);
\node at (2.6,-0.25) {\scriptsize Context 70\%};
\draw[rounded corners=8pt,fill=yellow!20] (9.6,0.0) rectangle (12.8,-0.5);
\node at (11.2,-0.25) {\scriptsize New lines since 30\%};

\draw[rounded corners=2pt,fill=white] (0.2,-1.0) rectangle (12.6,-2.4);
\node[anchor=west,font=\bfseries] at (0.35,-1.35) {Patient health record};
\node[anchor=west,align=left] at (0.35,-1.80) {Age 54, F, HTN, T2DM; meds: metformin, lisinopril; prior migraine; BMI 32.};

\draw[rounded corners=2pt,fill=white] (0.2,-2.8) rectangle (12.6,-4.6);
\node[anchor=west,font=\bfseries] at (0.35,-3.15) {Physician-patient dialogue (excerpt)};
\node[anchor=west,align=left] at (0.35,-3.60) {\scriptsize \emph{Excerpt continues; new lines at this context highlighted.}};

\draw[rounded corners=10pt,fill=blue!10] (0.2,-5.1) rectangle (3.9,-5.7);
\node[anchor=west] at (0.35,-5.40) {Question A};
\draw[rounded corners=10pt,fill=blue!10] (4.2,-5.1) rectangle (7.9,-5.7);
\node[anchor=west] at (4.35,-5.40) {Question B};
\draw[rounded corners=10pt,fill=blue!10] (8.2,-5.1) rectangle (11.9,-5.7);
\node[anchor=west] at (8.35,-5.40) {Question C};

\foreach \i/\label in {0/Guideline Navigation,1/Thought Alignment,2/Usefulness} {
  \pgfmathsetmacro{\yy}{-6.2 - 1.0*\i}
  \node[anchor=west] at (0.2,\yy) {\scriptsize \label};
  \foreach \j in {1,...,7} {
    \draw (3.7+0.6*\j,\yy) circle (0.14);
    \node[font=\scriptsize] at (3.7+0.6*\j,\yy-0.35) {\j};
  }
}

\draw[rounded corners=2pt,fill=white] (0.2,-8.8) rectangle (12.6,-10.0);
\node[anchor=west,font=\bfseries] at (0.35,-9.10) {Select the single most useful question};

\def\optY{-9.60} 

\foreach \x/\lab in {1/A,3/B,5/C} {
  \draw (\x,\optY) circle (0.14);
  \node[anchor=west] at (\x+0.25,\optY-0.03) {\scriptsize \lab};
}

\draw (7,\optY) circle (0.14);
\node[anchor=west] at (7.25,\optY-0.03) {\scriptsize No question needed};

\draw (11,\optY) circle (0.14);
\node[anchor=west] at (11.25,\optY-0.03) {\scriptsize Other};

\end{tikzpicture}
}
\caption{Sample questionnaire page at a given context level with three candidate questions, Likert items, and a best question choice.}
\label{fig:eval-questionnaire}
\end{figure*}

\section{LLM-as-judge Results using Open-source Model MedGemma-27B}\label{apd:medgemma}

We additionally report results from the state-of-the-art open-source medical model MedGemma-27B (text-only) \citep{sellergren2025medgemma} using the same automatic evaluation pipeline used for Gemini 2.5 Pro in Table~\ref{tab:medgemma}. Due to resource constraints, we do not include a full human evaluation of MedGemma. Therefore, Table~\ref{tab:medgemma} is not to benchmark model performance, but to demonstrate that our question generation pipeline is model-agnostic and can be reproduced with open-source or emerging models by the community.

\begin{table}[ht]
\centering
\caption{LLM-as-judge evaluation results across five dimensions by MedGemma-27B (text-only).}
\begin{tabular}{lcc}
\toprule
Metric & Zero-shot & Multi-stage \\
\midrule
Average              & 6.36 & 6.08 \\
Relevance            & 6.6  & 6.2  \\
Guideline Navigation & 6.8  & 5.8  \\
Thought Alignment    & 6.4  & 6.2  \\
Non-Redundancy       & 5.6  & 6.2  \\
Usefulness           & 6.4  & 6.0  \\
\bottomrule
\end{tabular}
\label{tab:medgemma}
\end{table}

\section{Correlation Analysis}\label{apd:correlation}

To quantify the alignment between automatic LLM-based scoring and human evaluation, we compute Spearman correlations between the LLM-as-judge scores and human physician ratings for all five metrics in Table \ref{tab:spearman_per_model}. The correlations are consistently weak ($\rho$ close to 0) and mostly non-significant across both the zero-shot baseline (M2) and the multi-stage pipeline (M1). When averaged across all stages and methods (Table \ref{tab:spearman_avg}), the correlations remain near zero. It indicates that LLMs do not reliably approximate clinician judgment in this setting. This analysis further supports our decision to treat LLM-as-judge only as a supplementary signal rather than a replacement for human evaluation.

\begin{table}[t]
\centering
\caption{Spearman correlation between LLM-as-judge scores and human physician ratings across five evaluation metrics, reported separately for the zero-shot baseline (M2) and the proposed multi-stage pipeline (M1).}
\begin{tabular}{lrr}
\toprule
Metric & $\rho$ & $p$ \\
\midrule
Relevance-M1       &  0.0718  & 0.2677 \\
Relevance-M2       &  0.0519  & 0.4237 \\
Guideline Navigation-M1   &  0.1190  & 0.0657 \\
Guideline Navigation-M2   &  0.0212  & 0.7439 \\
Thought Alignment-M1   &  0.0804  & 0.2145 \\
Thought Alignment-M2   & -0.0335  & 0.6055 \\
Non-Redundancy-M1   & -0.0060  & 0.9262 \\
Non-Redundancy-M2   & -0.0363  & 0.5759 \\
Usefulness-M1      &  0.0722  & 0.2652 \\
Usefulness-M2      & -0.0401  & 0.5367 \\
\bottomrule
\end{tabular}
\label{tab:spearman_per_model}
\end{table}

\begin{table}[t]
\centering
\caption{Average Spearman correlation between LLM-as-judge scores and physician ratings across five evaluation metrics, computed by aggregating correlations over all dialogue stages and both generation methods.}
\begin{tabular}{lr}
\toprule
Metric & Average Spearman Correlation \\
\midrule
Relevance            & -0.013329 \\
Guideline Navigation & -0.025142 \\
Thought Alignment    &  0.042234 \\
Non-Redundancy       & -0.048712 \\
Usefulness           &  0.014104 \\
\bottomrule
\end{tabular}
\label{tab:spearman_avg}
\end{table}

\end{document}